%% file: main.tex
\documentclass[11pt]{article}

\usepackage[utf8]{inputenc}
\usepackage{deauthor}
\usepackage{times,graphicx}

\usepackage{todonotes}
\usepackage{pifont}

\usepackage{multirow}
\usepackage{booktabs}
\usepackage{caption,subcaption}
\usepackage{graphicx}
\usepackage{xspace}
\usepackage{url}
\usepackage{amsmath,amssymb,amsfonts}
\usepackage{algorithm}
\usepackage{algorithmicx}
\usepackage[noend]{algpseudocode}

\title{FedCLIP: Fast Generalization and Personalization for CLIP in Federated Learning}

\author{Wang Lu$^1$
\hspace{2em} Xixu Hu$^2$
\hspace{2em} Jindong Wang$^{3}$\thanks{Corresponding author: Jindong Wang: jindong.wang@microsoft.com.}
\hspace{2em} Xing Xie$^{3}$\\
$^{1}$ Chinese Academy of Sciences, Beijing, China \\
$^{2}$ City University of Hong Kong, Hong Kong\\
$^{3}$ Microsoft Research Asia, Beijing, China \\
\texttt{\small luwang@ict.ac.cn, xixuhu2-c@my.cityu.edu.hk, \{jindong.wang, xingx\}@microsoft.com}\\
}

\newcommand{\algorithmname}{Algorithm}
\newcommand{\equationname}{Eq.}

\newcommand{\sectionname}{Sec.}

\usepackage{paralist}

\newcommand{\method}{FedCLIP\xspace}
\newcommand{\mecom}{AttAI\xspace}

\usepackage{todonotes}
\usepackage{xcolor}

\begin{document}

\maketitle
\begin{abstract}
Federated learning (FL) has emerged as a new paradigm for privacy-preserving computation in recent years. 
Unfortunately, FL faces two critical challenges that hinder its actual performance: data distribution heterogeneity and high resource costs brought by large foundation models.
Specifically, the non-IID data in different clients make existing FL algorithms hard to converge while the high resource costs, including computational and communication costs that increase the deployment difficulty in real-world scenarios. 
In this paper, we propose an effective yet simple method, named \method, to achieve fast generalization and personalization for CLIP in federated learning.
Concretely, we design an attention-based adapter for the large model, CLIP, and the rest operations merely depend on adapters. 
Lightweight adapters can make the most use of pretrained model information and ensure models be adaptive for clients in specific tasks.
Simultaneously, small-scale operations can mitigate the computational burden and communication burden caused by large models.
Extensive experiments are conducted on three datasets with distribution shifts.
Qualitative and quantitative results demonstrate that \method significantly outperforms other baselines ($\mathbf{9}\%$ overall improvements on PACS) and effectively reduces computational and communication costs ($\textbf{283x}$ faster than FedAVG). 
Our code will be available at: \url{https://github.com/microsoft/PersonalizedFL}.
\end{abstract}

\section{Introduction}

The success of machine learning, especially deep learning, is inseparable from a large amount of data.
However, data, as an important resource, usually scatter across different individuals or organizations.
In recent years, people pay more attention to data privacy and security and some organizations even enact relevant regulations and laws, e.g. The EU general data protection regulation (GDPR)~\cite{voigt2017eu} and China's cyber power~\cite{inkster2018china}.
Under this circumstance, direct raw data communication can be impossible in reality, making traditional data-centric machine learning paradigms unlikely to work.
To cope with this challenge, federated learning (FL)~\cite{yang2019federated} emerges as a new distributed machine learning paradigm and has been widely adopted in various applications.

Federated learning makes it possible to perform model aggregation without directly accessing the raw user data from different clients.
One of the earliest works in FL is called FedAVG~\cite{mcmahan2017communication} which aggregates distributed information using a simple and powerful averaging algorithm.
FedAVG mainly contains four steps, including training local models with local data, uploading local models to the server, aggregating models in the server, and distributing the aggregated model to each individual or organization.
These four steps are executed for multiple rounds for better information aggregation.
FedAVG can ensure that raw data does not leave the local client and thus protect data privacy and security.
Due to its simplicity and great performance, FedAVG quickly became popular in many areas~\cite{li2020review,rodriguez2023survey,banabilah2022federated}.

\begin{figure}[t]
	\centering
	\begin{subfigure}[b]{0.45\textwidth}
		\centering
		\includegraphics[height=0.8\textwidth]{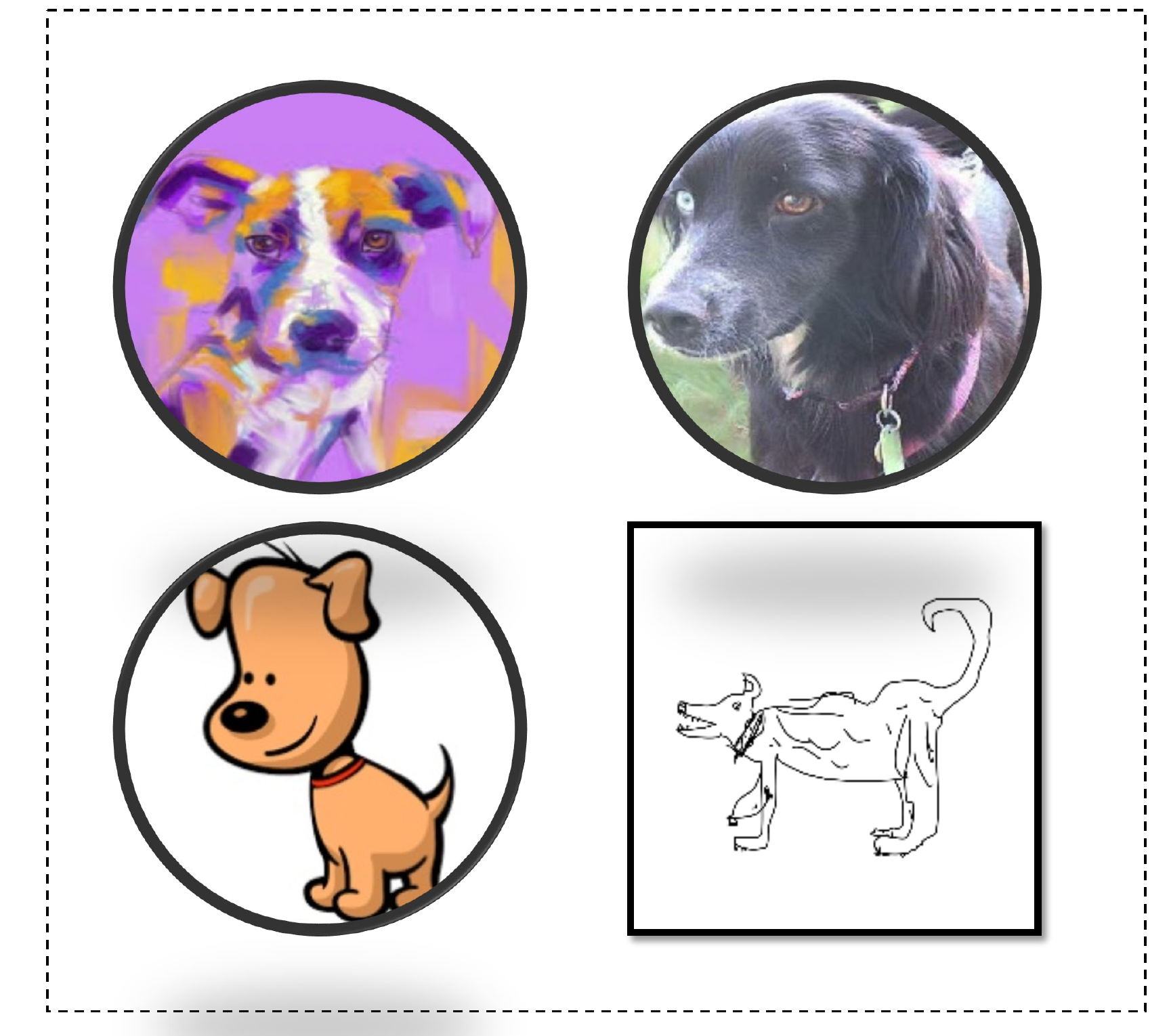}
		\caption{Data distribution shifts}
		\label{fig:issue1}
	\end{subfigure}
	\begin{subfigure}[b]{0.45\textwidth}
		\centering
		\includegraphics[height=0.8\textwidth]{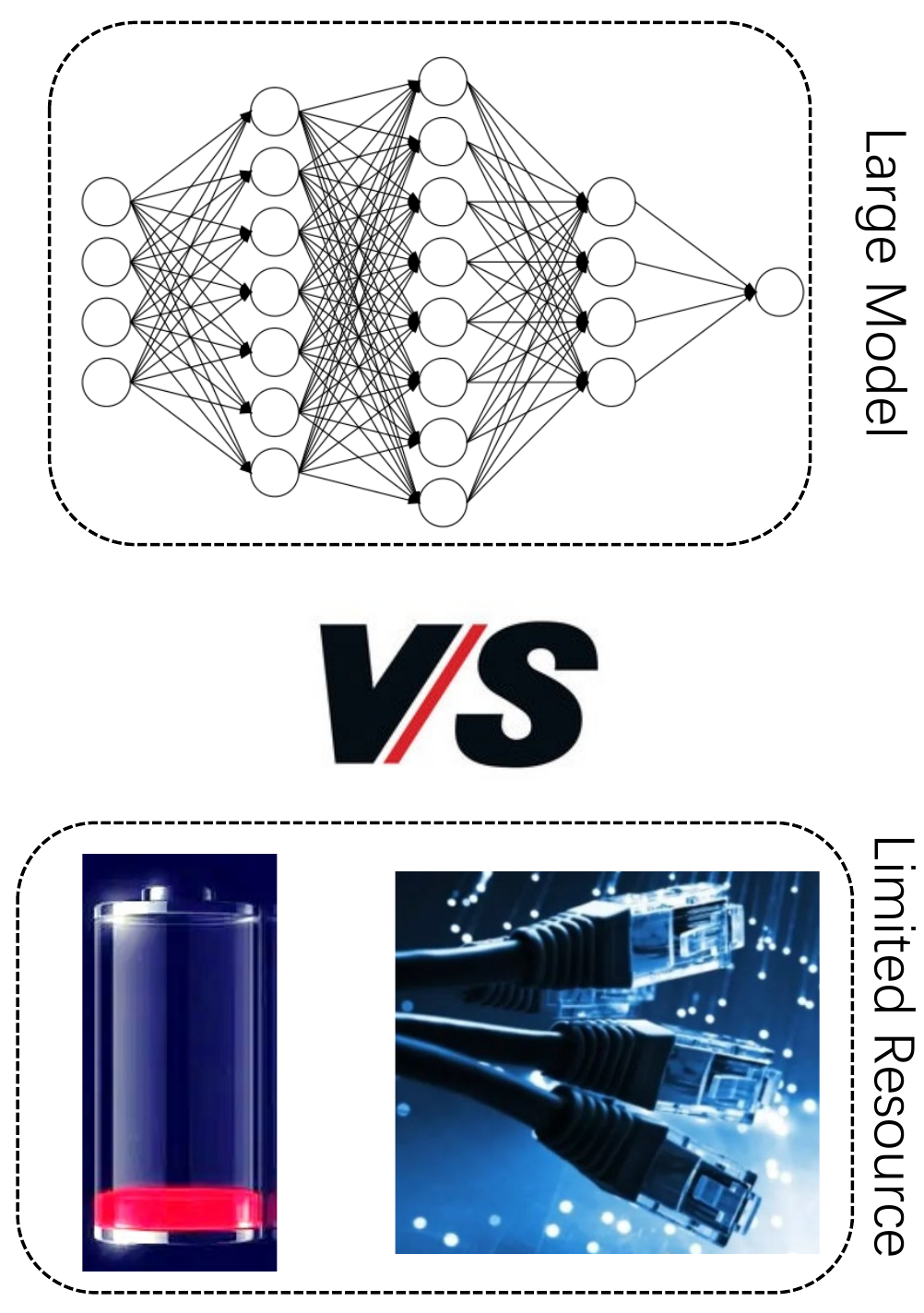}
		\caption{Huge resource demands}
		\label{fig:issue2}
	\end{subfigure}
	\caption{Existing issues in federated learning. In \figurename~\ref{fig:issue1}, circles denote participated clients while squares denote unseen targets.}
	\label{fig:issues}
\end{figure}

In this paper, we are specially interested in federated learning under the \emph{large foundation models} era~\cite{bommasani2021opportunities}.
Foundation models, as suggested by the name, have become increasingly popular in different machine learning tasks, such as Vision Transformer in computer vision~\cite{yuan2021tokens} and the GPT series in natural language processing~\cite{radford2018improving}. 
Since these models are extremely large, e.g., GPT-3~\cite{brown2020language} has 175 billion parameters, our key question is: \emph{how to perform effective and efficient federated learning using these large models?}

Specifically, two critical research challenges arise in this situation: \emph{data distribution shifts} and \emph{huge resource demands}.
On the one hand, data distribution shifts widely exist in the real world, e.g. figures shown in \figurename~\ref{fig:issue1}.
When meeting heterogeneous data, common federated learning methods can suffer from slow convergence and low accuracy due to inconsistent optimization directions, local optima, or some other factors~\cite{gao2022feddc}.
A qualified FL model can cope with both various clients and unseen targets, i.e. personalization and generalization.
On the other hand, huge resource demands of increasingly popular large models lead to conflicts with realistically constrained resources, as shown in \figurename~\ref{fig:issue2}.
In addition to high computational costs, communication cost is also a critical metric in federated learning.
For instance, the CLIP~\cite{radford2021learning} model based on VIT-B/32 contains more than $10^8$ trainable parameters and most existing networks cannot afford to transmit it quickly.
Achieving fast generalization and personalization with minimal resource costs is an urgent issue to be addressed.

Some existing work tried to address the issues mentioned above~\cite{lu2022personalized, Honglinyuan2022, guo2022promptfl}.
FedAP~\cite{lu2022personalized} attempted to learn the similarity among clients and then leveraged the learned similarity matrix to guide aggregation.
FedAP could achieve acceptable personalization results but it ignored generalization.
Another paper~\cite{Honglinyuan2022} discussed two gaps, including the out-of-sample gap and the participation gap.
These two gaps correspond to goals of generalization and personalization respectively.
This paper performed extensive empirical studies to analyze these issues but it did not offer a possible solution for large models.
PromptFL~\cite{guo2022promptfl} only updated the prompts instead of the whole model to accelerate the whole process.
However, clients still require large amounts of computation and PromptFL is not designed for personalization and generalization.

In this paper, we propose \method to achieve fast generalization and personalization for CLIP in federated learning. 
Since larger pretrained models, e.g. CLIP, have contained enough prior information, our goal is to find where we should focus in specific tasks.
The core part of \method is \mecom, an attention-based adapter for the image encoder in CLIP.
Instead of finetuning whole networks, \mecom directly utilizes fixed features extracted by pretrained models and explores where \method should pay attention to for specific tasks.
Simply training \mecom can ensure \method preserving prior information as much as possible while it allows models adapted for specific tasks. 
Through \mecom, \method does not rely on pretrained models anymore once obtaining diversified and robust features and thus \method can save large amounts of computational costs and communication costs.
Therefore, \method is extensible and can be deployed to many applications.


Our contributions are as follows.
\begin{enumerate}
    \item We propose \method, a fast generalization and personalization learning method for CLIP in federated learning. It can achieve personalization for participating clients and its remarkable generalization ability can attract new clients.
    \item Extensive experiments on three public image benchmarks demonstrate that \method can have achieved personalization and generalization performance at the same time ($\textbf{9}\%$ overall improvements on PACS). More importantly, \method reduces the number of trainable parameters thus saving communication costs and computational costs (\textbf{283x} faster than FedAVG).
    \item \method is extensible and can be applied in many real applications, which means it can work well in many circumstances. We can even embed it in some other architectures, e.g. BERT~\cite{tenney2019bert} and ViT~\cite{han2022survey}.
    Our code will be available at: \url{https://github.com/microsoft/PersonalizedFL}.
\end{enumerate}

The remainder of this paper is organized as follows.
In \sectionname~\ref{sec:relw}, we introduce related work.
And then we elaborate on the proposed method in \sectionname~\ref{sec:method}.
Extensive experiments are reported and analyzed in \sectionname~\ref{sec:exp}.
Finally, we conclude the paper and provide possible future work in \sectionname~\ref{sec:concl}.

\section{Related Work}
\label{sec:relw}
\input{relatework}

\section{Method}
\label{sec:method}
\subsection{Problem Formulation}
In a generalization and personalization federated learning setting, $N$ different clients, denote as $\{C_1, C_2, \cdots, C_N\}$, participate in exchanging information and they have data, denoted as $\{ \mathcal{D}_1, \mathcal{D}_2, \cdots, \mathcal{D}_N \}$ with different distributions, which means $P(\mathcal{D}_i) \neq P(\mathcal{D}_j)$. 
In this paper, we only focus on homogeneous data with the same input space and output space, i.e. $\mathcal{X}_i = \mathcal{X}_j, \mathcal{Y}_i = \mathcal{Y}_j, \forall i\neq j$.
Each dataset, $\mathcal{D}_i = \{ (\mathbf{x}_{i,j}, y_{i,j}) \}_{j=1}^{n_i}$, consists of three parts, a training dataset $\mathcal{D}_i^{train} = \{ (\mathbf{x}_{i,j}^{train}, y_{i,j}^{train}) \}_{j=1}^{n_i^{train}}$, a validation dataset $\mathcal{D}_i^{valid} = \{ (\mathbf{x}_{i,j}^{valid}, y_{i,j}^{valid}) \}_{j=1}^{n_i^{valid}}$ and a test dataset $\mathcal{D}_i^{test} = \{ (\mathbf{x}_{i,j}^{test}, y_{i,j}^{test}) \}_{j=1}^{n_i^{test}}$. 
Three sub-datasets in each client have no overlap and $n_i = n_i^{train} + n_i^{valid} + n_i^{test}, \mathcal{D}_i = \mathcal{D}_i^{train} \cup \mathcal{D}_i^{valid} \cup \mathcal{D}_i^{test}$.
Our goal is to aggregate all clients' information with preserving data privacy and security and learn a good model $f$ for each client $\mathcal{D}_i$:
\begin{equation}
    \min_{f} \frac{1}{N} \sum_{i=1}^N \frac{1}{n_{i}^{test}} \sum_{j=1}^{n_i^{test}} \ell(f(\mathbf{x}_{i,j}^{test}), y_{i,j}^{test}),
    \label{eqa:goal1}
\end{equation}
where $\ell$ is a loss function.
Moreover, for generalization, we assume that there exist $M$ different clients, denote as $\{F_1, F_2, \cdots, F_M\}$, with data $\{ \mathcal{D}_1^F =\{ (\mathbf{x}_{i,j}, y_{i,j}) \}_{j=1}^{m_1},  \mathcal{D}_2^F= \{ (\mathbf{x}_{i,j}, y_{i,j}) \}_{j=1}^{m_2}, \cdots, \mathcal{D}_N^F=\{ (\mathbf{x}_{i,j}, y_{i,j}) \}_{j=1}^{m_M} \}$.
These $M$ clients do not participate in training, and we hope $f$ can also be able to perform well on these clients.
\begin{equation}
    \min_{f} \frac{1}{M} \sum_{i=1}^M \frac{1}{m_{i}} \sum_{j=1}^{m_i} \ell(f(\mathbf{x}_{i,j}), y_{i,j}),
    \label{eqa:goal2}
\end{equation}

\subsection{Preliminaries}

\paragraph{CLIP}

CLIP, Contrastive Language Image Pre-training, is an efficient and scalable method of learning~\cite{radford2021learning}.
To compensate for the problems caused by the amount of data and model parameters, it trained a large model with over $4\times 10^8$ pairs of data.
With help of natural language supervision, CLIP can better understand concepts of visual images and better learn the semantic connections behind images.
Usually, CLIP models contain more information and they might be more robust.

A simple CLIP model regularly contains two parts, an image encoder $f^I$ and a text encoder $f^T$.
In common models, labels are frequently represented as numbers or one-hot vectors.
For CLIP, to better utilize semantic information, these labels are often transformed into sentences, e.g. 'A photo of dogs'.
And then text feature vectors, $\mathbf{T}$ are extracted from these sentences via $f^T$.
Concurrently, images are encoded into visual feature vectors, $\mathbf{I}$, via $f^I$.
Cosine similarities between $\mathbf{T}$ and $\mathbf{I}$ are used to training and predicting.

\paragraph{FedAVG}
In FedAVG~\cite{mcmahan2017communication}, each client trains $f$ with local clients' data, and then parameters of updated models, $w_i$, are transmitted to the server.
The server typically aggregates the parameters according to \equationname~\ref{eq:fedagg},
\begin{equation}
    w^*=\sum_{i=1}^N \frac{n_i}{\sum_{j=1}^N n_j} w_i
    \label{eq:fedagg}
\end{equation}
After aggregation, $w^*$ is distributed. 
When $|w|$ is larger, the server cannot afford communication costs.

\subsection{\method}
To reduce computational costs and communications and make the most use of existing pretrained model information, we propose \method.
Pretrained models already have abilities to extract robust and diversified features.
Tuning whole networks with limited data can compromise the original ability of pretrained models.
What we need to do is to try our best to preserve useful prior knowledge and let it be used to a suitable extent for our task.
Besides, tuning large networks is impractical in federated learning due to limited resources in reality.
Therefore, instead of operating on the whole model, \method concentrates on a simple attention-based adapter for the image encoder, \mecom.

\begin{figure}[t]
	\centering
    \includegraphics[width=.9\textwidth]{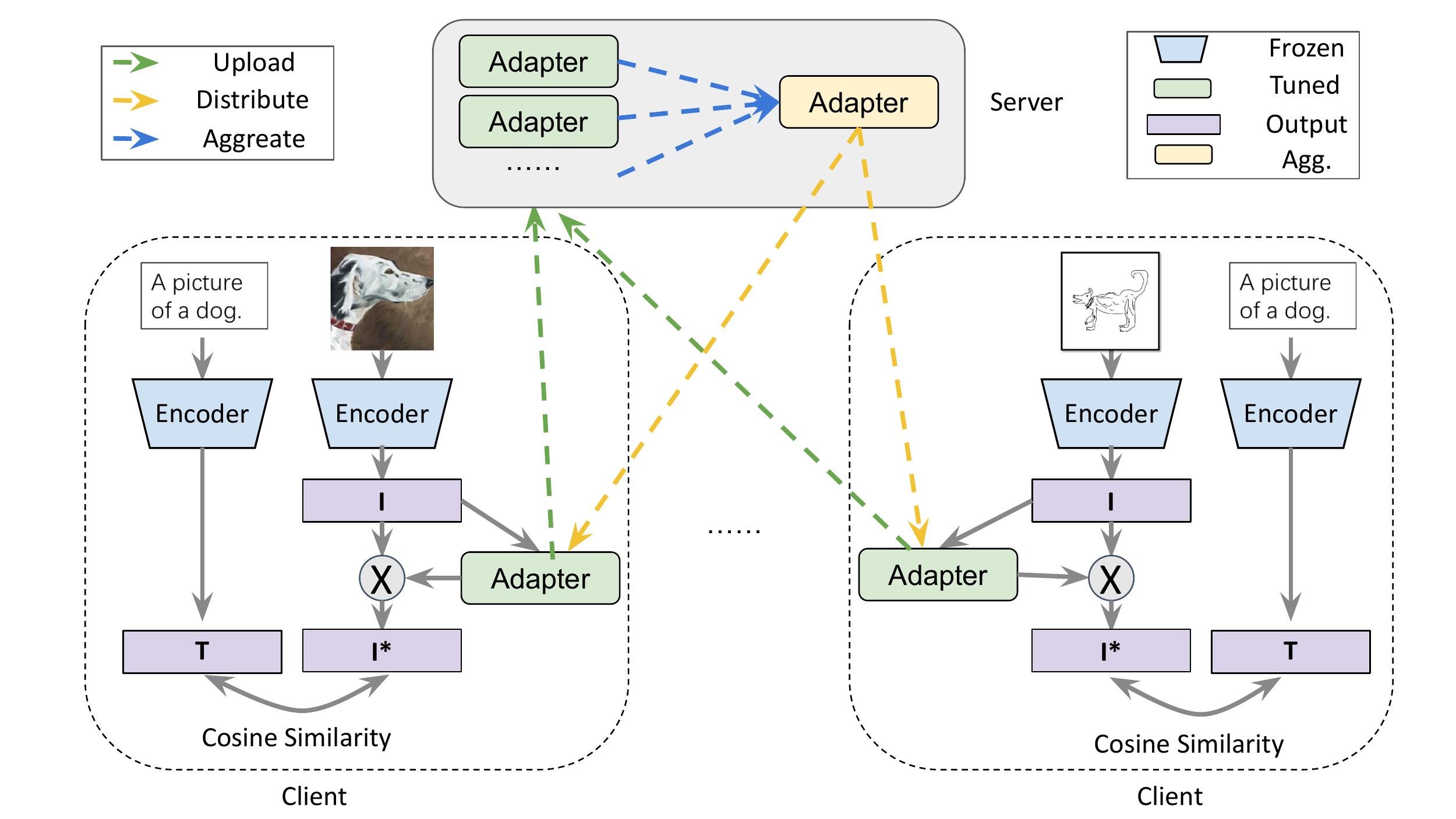}
	\caption{The framework of \method.} 
	\label{fig:frame1}
\end{figure}

\figurename~\ref{fig:frame1} gives the framework of \method.
As shown in \figurename~\ref{fig:frame1}, our method mainly contains four steps.

\begin{enumerate}
    \item For Client $i$, we utilize a pretrained CLIP model to extract features of data, denoted as $T_i$ and $I_i$.
    \item In each client, we utilize $\mathcal{D}^{train}_i$ to train the corresponding adapter, $g_i$. And then we upload $\{g_i\}_{i=1}^N$ to the server.
    \item In the server, the parameters of all $g_i$ are weighted averaged and we can obtain $g^*$. The server then distributes $g^*$ to each client and updates the parameters of each $g_i$.
    \item Repeat Step 2 and Step 3 until convergence or reaching maximum rounds.
\end{enumerate}

In step 1, we utilize the pretrained CLIP model to extract features. We consider the pretrained model is so powerful that we do not need to explore some other features.
For $(\mathbf{x},y)$, we can obtain corresponding features, 
\begin{equation}
\mathbf{I}=f^I(\mathbf{x}),
\mathbf{T}=f^T(y)
\end{equation}

What we need to do next is to identify which parts of features are suitable for our specific tasks.
Therefore, we introduce an attention-based adapter, $g$, to locate where we should concentrate on.
Particularly, we utilize one linear layer, Tahn activation function, one linear layer, and Softmax activation function to construct $g$.
The Softmax function is used to ensure our final outputs ranging from $0$ to $1$.
Once we obtain the attention vector $att=g(\mathbf{I})$, we utilize it to update the visual feature via a dot multiply operation, 
\begin{equation}
    \mathbf{I}^*= g(\mathbf{I})\cdot \mathbf{I}.
\end{equation}

Then, similar to \cite{radford2021learning}, we normalize $\mathbf{I}^*$ and $\mathbf{T}$ to compute the final logits.

\begin{align}
    \title{\mathbf{I}} = \frac{\mathbf{I}^*}{|I^*|}, & \title{\mathbf{T}} = \frac{\mathbf{T}}{|\mathbf{I}|},\\
    \hat{\mathbf{I}}= s \times \title{\mathbf{I}} \times \title{\mathbf{T}}^T,&  \hat{\mathbf{T}}=\hat{\mathbf{I}}^T.
\end{align}
where $s$ is a scale parameter.

Now, we can utilize the ground truth, a vector $\tilde{\mathbf{y}}=[0,1,2,3,\cdots, B]$
\begin{equation}
\begin{split}
        \ell_{cls}^I= \ell(\hat{\mathbf{I}},\tilde{\mathbf{y}}),\\
    \ell_{cls}^T= \ell(\hat{\mathbf{T}},\tilde{\mathbf{y}}),
\end{split}
\end{equation}
where $\ell$ is CrossEntropy loss~\cite{zhang2018generalized} while $B$ is the number of images in a batch.

We only exchange parameters of adapters, $w^g$, and therefore in the server, we replace \equationname~\ref{eq:fedagg} with \equationname~\ref{eq:ada}.

\begin{equation}
    w^{g,*}=\sum_{i=1}^N \frac{n_i}{\sum_{j=1}^N n_j} w_i^g.
    \label{eq:ada}
\end{equation}

Since $w^g$ contains substantially less amount of trainable parameters than $w$, \method saves computational costs and communication costs. 
\subsection{Summary}
For clarity, we give a detailed description of \method in \algorithmname~\ref{alg:adafed}.
In Line 1, directly obtaining generalized and diversified features with fixed CLIP make it possible to utilize more prior knowledge of pretrained models.
In Line 2, with adapters, we can concentrate on valuable information and eliminate the influence of redundant information in specific tasks.
Rich prior knowledge and targeted attention make the ultimately extracted features more robust, effective, and adaptable, resulting in our method having good generalization and personalization capabilities.
From Line 2 to Line 5, performing computation and transmission merely with adapters can save a lot of resources and ensure the efficiency of our method.

\begin{algorithm}[htb]
\caption{\method}
\label{alg:adafed}
\textbf{Input}: $N$ clients' datasets $\{\mathcal{D}_i\}_{i=1}^N$, a pretained CLIP model consist of an image encoder, $f^I$, and a text encoder, $f^T$\\
\textbf{Output}: An adapter $g$ 
\begin{algorithmic}[1] 
\State For client $i$, computer the corresponding features $I_i=f^I(\mathbf{X}_i), T_i=f^T(\mathbf{Y}_i)$
\State For client $i$, train the local adapter, $g_i$, according to \equationname 5 to \equationname 9
\State Send the current adapter $g_i$ to the server
\State Aggregate adapters' parameters via \equationname~\ref{eq:ada} and obtain $w^{g*}$
\State Transmit $w^{g*}$ to each client
\State Repeat steps $2 \sim 5$ until convergence
\end{algorithmic}
\end{algorithm}

\subsection{Discussion}
Adapter is a common technique in transfer learning~\cite{wenxinhou22}.
It is at a small scale and has a plug-and-play implementation.
In this paper, we mainly focus on adaptations to image encoders.
Actually, we also can add adapters to text encoders.
We can even change the inputs of text encoders to incorporate more semantic information.

\section{Experiments}
\label{sec:exp}

In this section, we extensively evaluate \method in three common visual image classification benchmarks.

\subsection{Datasets}
\paragraph{PACS} PACS~\cite{li2017deeper} is a popular object classification benchmark. 
It is composed of four sub-datasets, including photo, art-painting, cartoon, and sketch.
There exist $9,991$ images in total and the dataset contains $7$ classes, including dog, elephant, giraffe, guitar, horse, house, and person.
Large discrepancies in image styles widely exist among different sub-datasets.
In this paper, we view each sub-dataset as a client. 
We choose three sub-datasets as participated clients while the rest served as the target client to evaluate generalization ability.
For each participated client, we split the corresponding sub-dataset into three parts, $60\%$ for training, $20\%$ for validation, and the rest $20\%$ for testing.
Validation parts of data are used for model selection.

\paragraph{VLCS} VLCS~\cite{fang2013unbiased} is another widely accepted public image classification benchmark.
It also consists of four sub-datasets (VOC2007, LabelMe, Caltech10, and SUN09).
It contains $10,729$ instances with 5 classes.
Feature shifts exist generally among different sub-datasets.
Similar to PACS, four sub-datasets correspond to four clients.
Three sub-datasets play the roles of participants while the rest one act as an upcoming client.

\paragraph{Office-Home} Office-Home~\cite{venkateswara2017deep} is a larger image classification benchmark, which contains $65$ classes.
Office-Home comprises four sub-datasets (Art, Clipart, Product, and Real\_World) with about $15,500$ images.
The feature shifts from Office-Home mainly come from image styles and viewpoints, but they are much smaller than PACS.
We assess methods on Office-Home in a similar manner to PACS.

\subsection{Implementation Details and Comparison Methods}
For these three common image classification benchmarks, we use the CLIP pre-trained model with ViT-B/32~\cite{dosovitskiyimage} as the image encoder.
For model training, we utilize cross-entropy loss and Adam optimizer.
The learning rate is tuned from $5\times 10^{-5}$ to $5\times 10^{-3}$.
We set local update epochs as $E=1$ where $E$ means the number of training epochs in one round while we set the total communication round number as $R=200$.
Since, at each time, we set one sub-dataset as the target, i.e. upcoming client, there exist four tasks for each benchmark.
We run three trials to record the average results.
To better illustrate the function and necessity of using larger pretrained models, we also utilize a related small architecture, AlexNet~\cite{krizhevsky2012imagenet}, to perform some base federated learning methods.

We compare our method with two methods including a common federated learning method, FedAVG, and a method designed for non-iid data, FedProx.
\begin{enumerate}
    \item FedAVG~\cite{mcmahan2017communication}. The server aggregates all client models' parameters. FedAVG will aggregate networks with several layers for AlexNet while FedAVG will aggregate both image encoders and text encoders for CLIP.
    \item FedProx~\cite{li2018federated}. It adds a proximal term to FedAVG and allows the existence of slight differences between clients and the server.
\end{enumerate}

\subsection{Results}
\paragraph{Generalization Ability}

\input{tab-genera}
We first evaluate the generalization ability of each method via accuracy on clients that do not participate in training. 
\tablename~\ref{tab:my-table-ca} shows the generalization results for each task on PACS and Office-Home.
We have the following observations from these results.
1) Our method achieves the best generalization ability on average with remarkable improvements (about $14\%$ for PACS and about $12\%$ for Office-Home).
Moreover, our method achieves the best generalization ability in each task, which demonstrates the excellent generalization ability of our method.
2) Compared to methods with AlexNet as the backbone, methods with CLIP as the backbone can obtain better performance.
It demonstrates that large well-trained models can be able to bring better generalization.
3) Compared to methods with CLIP as the backbone, our method has a further improvement, which demonstrates that our method leverages prior knowledge better.

\paragraph{Personalization Ability}
\input{tab-person}
Then, we evaluate the personalization ability of each method via the accuracy on test data of each participating client.
\tablename~\ref{tab:my-table-person} shows the personalization results for each task on PACS and Office-Home.
We also have some insightful observations.
1)Although all clients share the same adapter in our method, our method still achieves the best average accuracy.
Moreover, \method almost achieves the best performance on each client for every task.
2) Compared to methods with AlexNet, corresponding methods with CLIP perform better overall.
For CLIP-based methods, results are quite sensitive to hyperparameters, e.g. learning rate.
And FedAVG has disappointing results on some specific clients.
3) Our method has the most use of prior knowledge since it achieves the stablest results.

\paragraph{Comprehensive Ability}
\input{tab-comp}
Finally, taking into account the performance of both personalization and generalization, we provide an overall performance in \tablename~\ref{tab:my-table-comp}.
Without a doubt, our method achieves the best overall performance with significant improvements (about $9\%$ for PACS and $8\%$ for Office-Home).
Compared to methods based on AlexNet, corresponding methods based on CLIP perform better.

\paragraph{More results on VLCS}
\input{tab-vlcs}
Due to space limitations, we only report comprehensive ability on VLCS.
As shown in \tablename~\ref{tab:my-table-comp-vlcs}, our method still achieves the best performance with improvements of over $10\%$.
Moreover, our method achieves the best in each task.
The results prove the superiority of our method again.

\subsection{Analysis}

\begin{figure}[t!]
	\centering
	\begin{subfigure}[b]{0.24\textwidth}
		\centering
		\includegraphics[width=\textwidth]{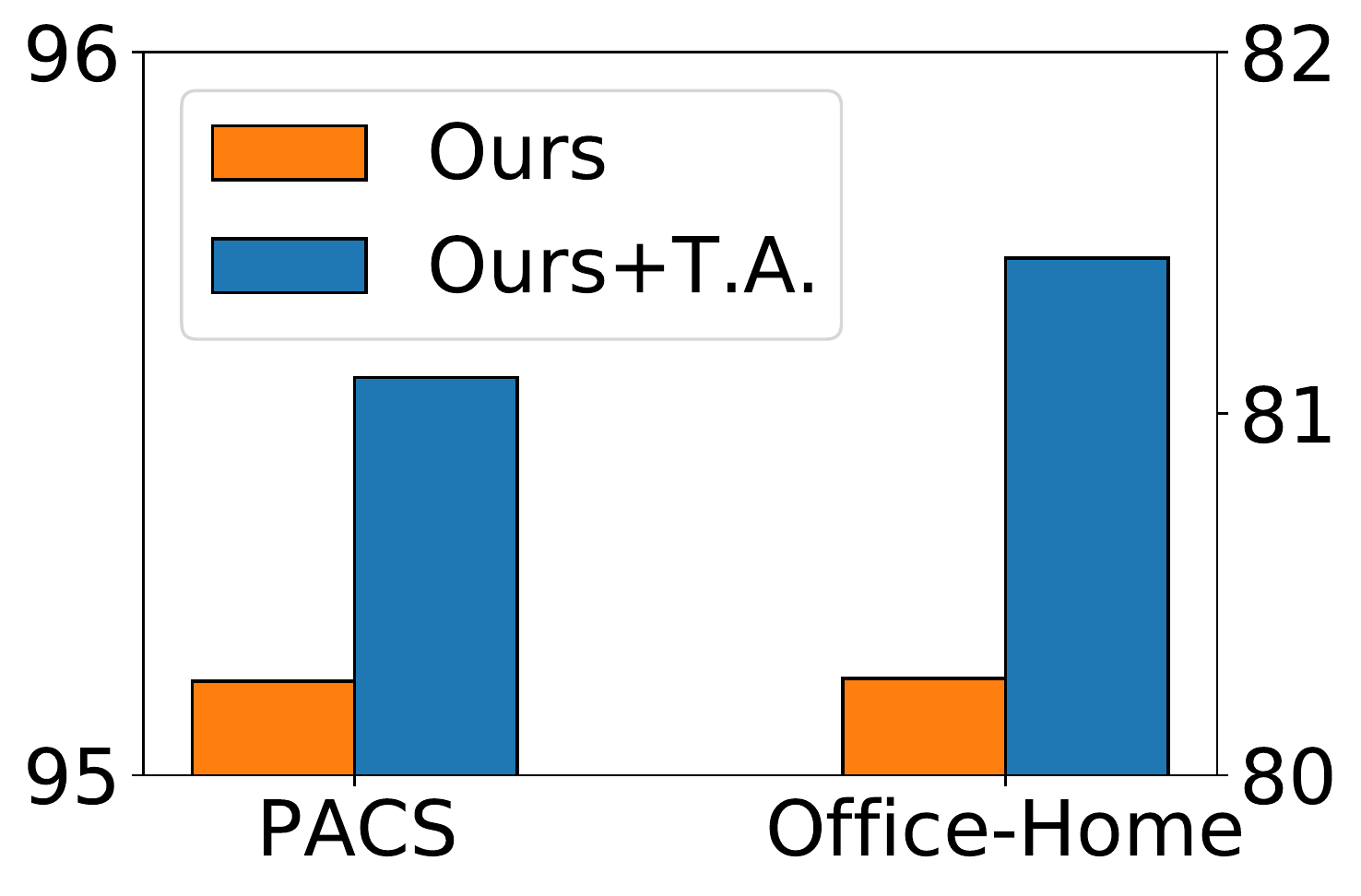}
		\caption{Adapter influence.}
		\label{fig:anly1}
	\end{subfigure}
    \begin{subfigure}[b]{0.24\textwidth}
		\centering
		\includegraphics[width=\textwidth]{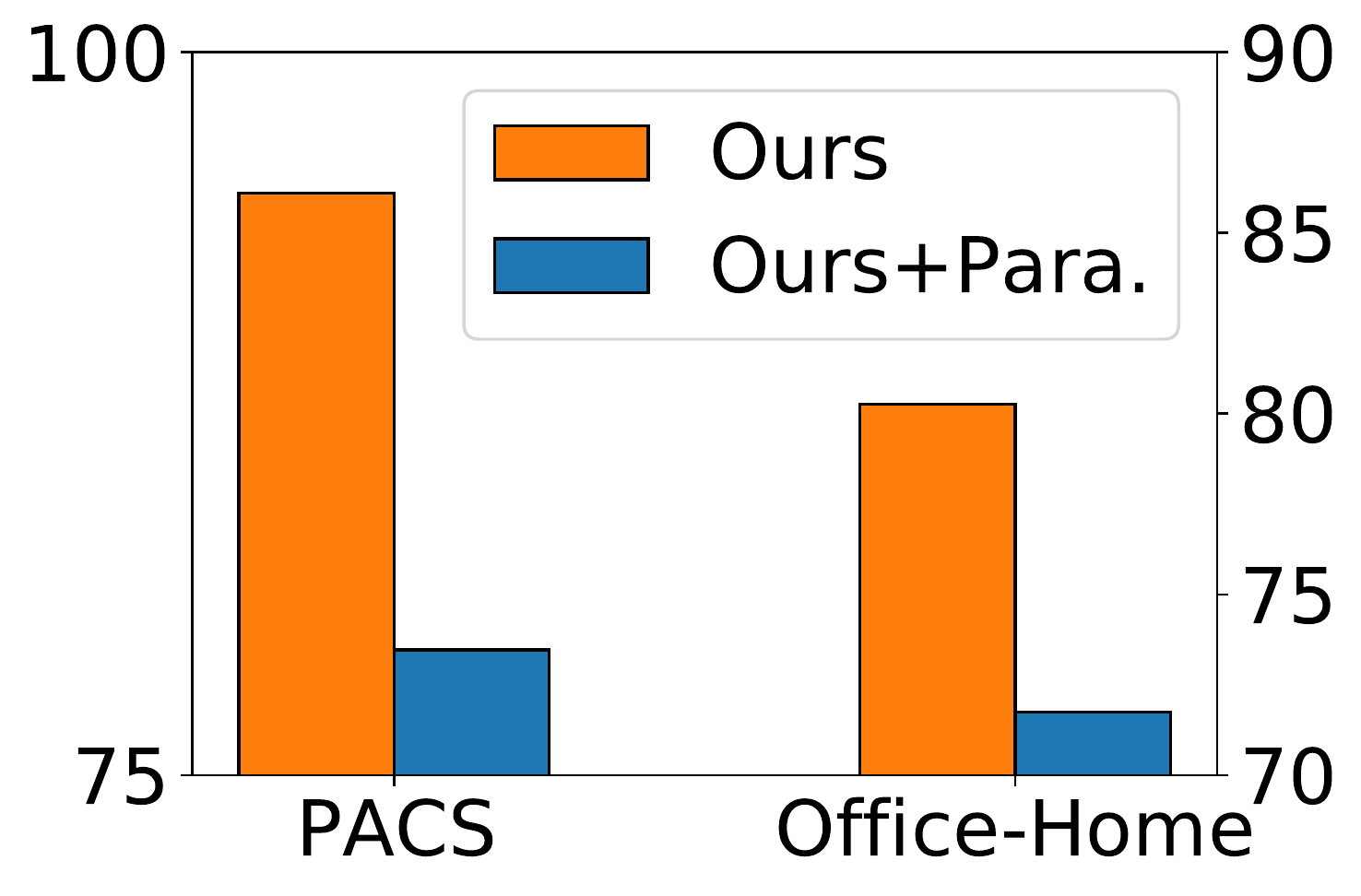}
		\caption{Training backbone.}
		\label{fig:anly2}
	\end{subfigure}
 	\begin{subfigure}[b]{0.24\textwidth}
		\centering
		\includegraphics[width=\textwidth]{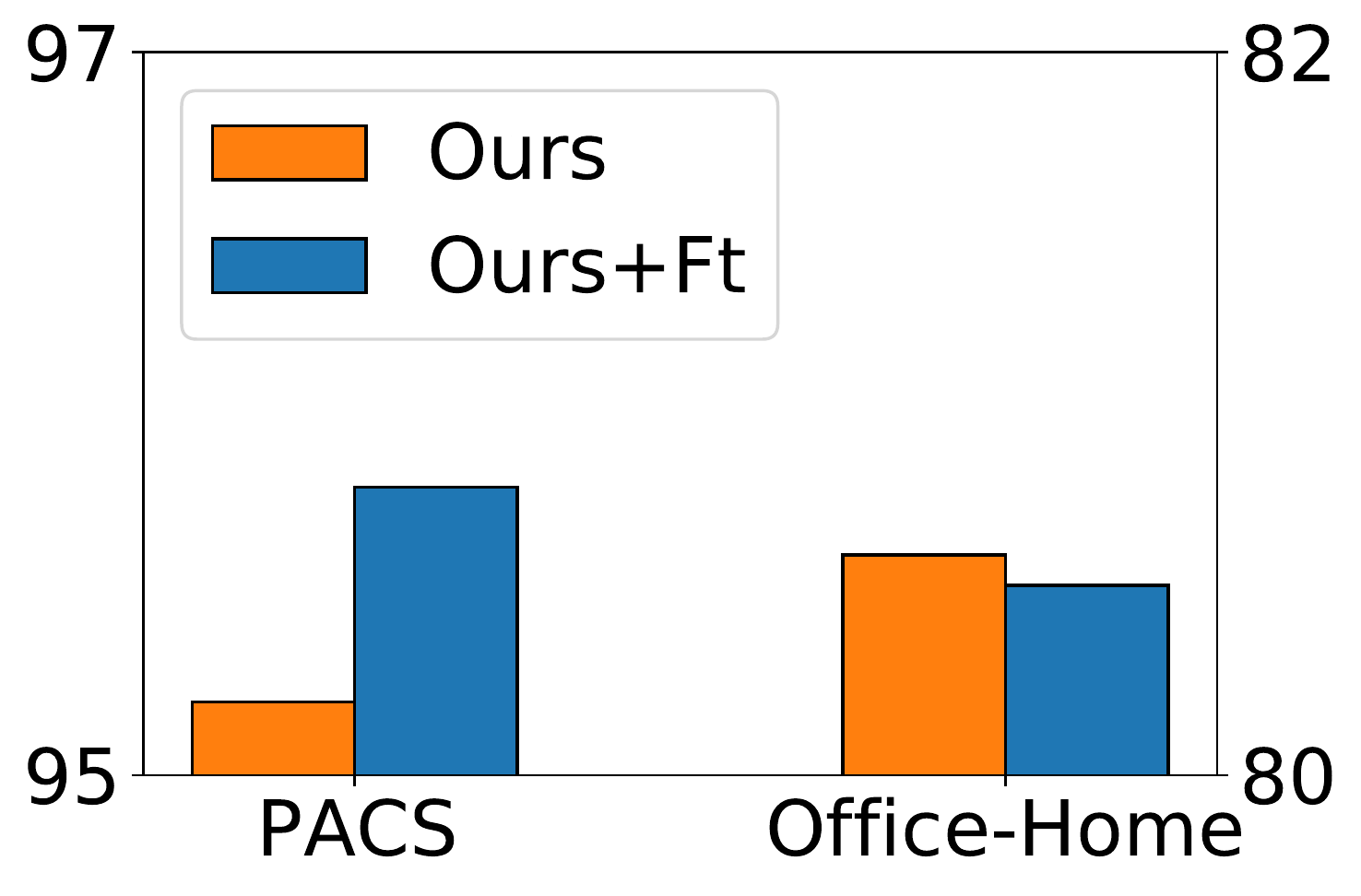}
		\caption{Finetune influence.}
		\label{fig:anly3}
	\end{subfigure}
 	\begin{subfigure}[b]{0.24\textwidth}
		\centering
		\includegraphics[width=\textwidth]{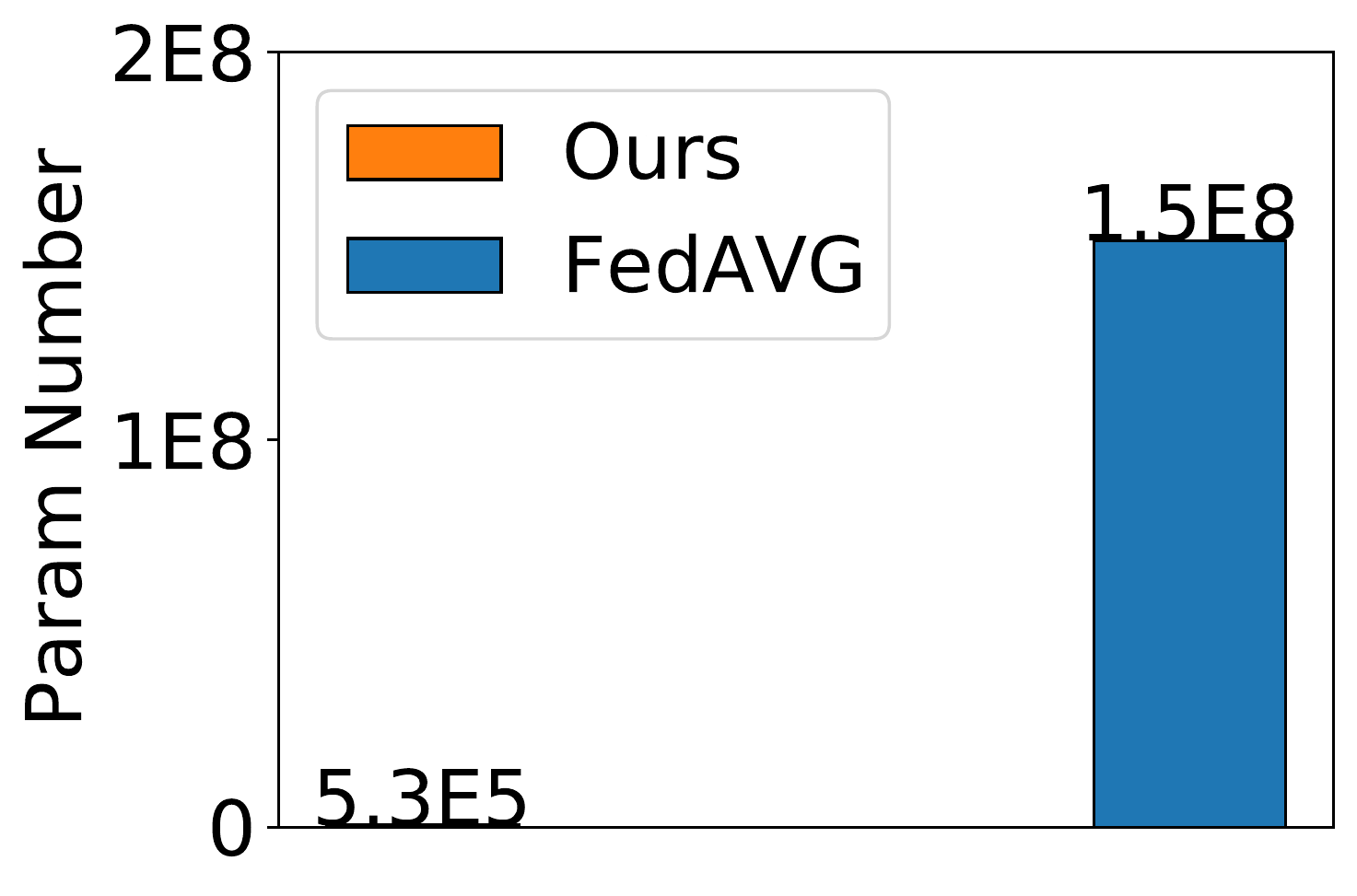}
		\caption{Parameter counts.}
		\label{fig:anly4}
	\end{subfigure}
	\caption{Analysis on PACS. }
	\label{fig:analysis}
\end{figure}

\paragraph{Can more adapters bring better performance?}
In our method, we only add one adapter to the image encoder.
We can add another adapter to the text encoder.
As shown in \figurename~\ref{fig:anly1}, adding more adapters brings slight improvements.
However, the improvements are so small that we need to assess whether it is necessary to do so since more adapters regularly mean more computational costs and more communication costs.

\paragraph{Can more trainable parameters bring better performance?}
If we train both adapters and the backbones, the results could be worse.
Since CLIP models have a wealth of good information, it is not suitable to change parameters with only a few data for a specific task.
Changes in CLIP with few data can destroy the feature extraction capabilities.
As shown in \figurename~\ref{fig:anly2}, we train more parameters but achieve worse performance.

\paragraph{Will finetuning bring better personalization?}
According to \cite{yu2020salvaging}, finetuning can be a useful technique for better personalization.
We also add experiments with finetune.
As shown in \figurename~\ref{fig:anly3}, finetune has no advance in personalization, which demonstrates that our method can be remarkable and robust when meeting non-iid.

\paragraph{Resource Cost Comparison}
The number of trainable parameters represents how many resources we need to cost in federated learning.
As shown in \figurename~\ref{fig:anly4}, our method merely has $5.3E5$ parameters while FedAVG with CLIP requires $1.5E8$ trainable parameters.
Common methods via training whole networks have $283$ times as many parameters as ours, which illustrates that our method is fast and resource-efficient.

\section{Conclusion and Future Work}
\label{sec:concl}
In this article, we propose \method, a fast generalization and personalization learning method for CLIP in federated learning.
\method designs an attention based adapter to replace updating the whole model.
Therefore, \method makes the most use of prior knowledge and saves computational costs and communication costs.
Comprehensive experiments have demonstrated the superiority of \method.
In the future, we plan to embed \method into more architectures and design more flexible adapters for different tasks.
We also plan to apply \method for heterogeneous architectures and more realistic applications.



\small
\bibliographystyle{abbrv}
\bibliography{ref.bib}

\end{document}

%% file: relatework.tex
\subsection{Challenges in Machine Learning}

Machine learning has achieved great success and gradually entered people's daily lives~\cite{sarker2021machine, paluru2021anam, wang2019deep}.
It has been applied to many fields, e.g. human activity recognition~\cite{lu2021cross}, face recognition~\cite{liu2022sphereface}, and healthcare~\cite{chen2022metafed}.
Successful machine learning applications, especially deep learning based applications, often require a large amount of data and lots of computational resources.
In most cases, a deluge of data and computing resources can lead to easy success, such as ChatGPT~\cite{van2023chatgpt}.
However, data and computation also mean money and resources.
In reality, it is impossible to aggregate all data together in some situations.
There seems to be a contradiction between the massive resource requirements of traditional methods and the limited real environment. 

Generalization is another challenging problem caused by data distribution shifts.
Its goal is to learn a generalized model with limited data and it expects that the learned model can work well on unseen targets with unknown distributions.
\cite{wang2022generalizing} gives a survey on domain generalization and first groups existing methods into three categories, including data manipulation~\cite{lu2022semantic}, representation learning~\cite{lu2022local}, and learning strategy~\cite{huang2020self}. 

\subsection{Federated Learning}
Data is often scatted everywhere and cannot be aggregated together due to some factors, such as laws and regulations~\cite{voigt2017eu} and the awakening of people's awareness of data security and privacy protection.
In such an environment, federated learning came into being~\cite{yang2019federated,roy2019braintorrent}.
According to \cite{yang2019federated}, federated learning can be grouped into three categories, including horizontal federated learning, vertical federated learning, and federated transfer learning.
Most deep learning based methods belong to horizontal federated learning and so is this paper.
For a more detailed introduction, please refer to the survey~\cite{liu2022distributed}.

FedAVG is a traditional horizontal federated learning method~\cite{mcmahan2017communication}.
Although it is simple, it was applied in many applications.
When meeting data distribution heterogeneity, FedAVG appeared powerless~\cite{sattler2019robust}.
And many researchers proposed various methods to solve the above problems.
FedProx~\cite{li2018federated} added a proximal regularized term to FedAVG and it allowed slight model gaps between clients and the server.
In FedBN~\cite{li2021fedbn}, the authors thought that parameters in batch normalization layers can represent data distribution, and keeping specific batch normalization layers for each client could make local models personalized.
Another latest method, FedAP~\cite{lu2022personalized}, learned the similarity between clients based on the statistics of the batch normalization layers while preserving the specificity of each client with different local batch normalization.
The above methods all achieve satisfactory results in their corresponding scenarios.
However, most of them focused on personalization and ignored generalization issues~\cite{chenbridging}.

Generalization in federated learning is a novel problem.
In recent two years, some papers tried to solve this problem.
\cite{Honglinyuan2022} first discussed the generalization in federated learning and it proposed a framework to disentangle performance gaps, including out-of-sample gaps and participation gaps.
FED-DRO~\cite{chenbridging} proposed a novel federated learning framework to explicitly decouple a model's dual duties with two prediction tasks and it mainly focused on label shifts.
Some other work tried to adapt existing domain generalization methods to generalization in federated learning~\cite{gupta2022fl,tenison2022gradient,qu2022generalized,caldarola2022improving}.
FL Games~\cite{gupta2022fl} utilized Nash equilibrium to learn causal features that were invariant across clients which is similar to Invariant Risk Minimization (IRM)~\cite{arjovsky2020invariant}.
FedSAM~\cite{qu2022generalized} proposed a general effective algorithm based on Sharpness Aware Minimization (SAM) local optimizer~\cite{foretsharpness}.
Although these methods can bring generalization, they were not designed for large models and could not make full use of knowledge brought by pretrained models.


\subsection{CLIP and Large Models}
From perceptron~\cite{gardner1998artificial} to AlexNet~\cite{alom2018history} to ResNet~\cite{he2016deep} to Vision transformer~\cite{yuan2021tokens} to CLIP~\cite{radford2021learning}, pretrained models have become larger and larger.
The importance of pretrained models has been increasing and pretrained models contain a growing amount of knowledge.
For specific applications, researchers usually choose suitable backbone models and then adopt some techniques, e.g. finetune~\cite{sun2019fine}, to slightly adapt pretrained models.
Since pretrained models are trained via a large amount of data, features extracted from them are often generalized and insightful.
Few works pay attention to large models in federated learning and high demands of computational costs and communication costs hinder the development of this field.
In this paper, we focus on CLIP in federated learning.

CLIP~\cite{radford2021learning} learned SOTA image representations from scratch on a dataset of 400 million(image, text) pairs collected from the internet.
The natural language was used to reference learned visual concepts.
It has been applied in many fields and demonstrated its superiority~\cite{ramesh2021zero, lee2023image}.
However, in federated learning, CLIP is still in its infancy.
PromptFL~\cite{guo2022promptfl} replaced the federated model training with the federated prompt training to simultaneously achieve efficient global aggregation and local training by exploiting the power of foundation models in a distributed way. 
However, it still requires certain computational costs and it is not designed for data distribution heterogeneity problems.
Moreover, it is hard to tune the hyperparameters for the prompt techniques in Transformer.
In this paper, we focus on fast personalization and generalization for CLIP.

%% file: tab-genera.tex
\begin{table}[t!]
\caption{Generalization accuracy. \textbf{Bold} means the best.}
\label{tab:my-table-ca}
\resizebox{\textwidth}{!}{
\begin{tabular}{llccccc|llccccc}
\toprule
\multicolumn{1}{l}{Dataset} & \multicolumn{6}{c|}{PACS}                                                                     & \multicolumn{7}{c}{Office-Home}                                                                                             \\
\multicolumn{1}{l}{Backbone} & Method  & A              & C              & P              & S              & AVG            & \multicolumn{1}{l}{Backbone} & Method  & A              & C              & P              & R              & AVG            \\ \midrule
\multirow{2}{*}{AlexNet}     & FedAVG  & 31.54          & 43.69          & 44.55          & 36.29          & 39.02          & \multirow{2}{*}{AlexNet}     & FedAVG  & 15.70          & 17.00          & 31.56          & 28.99          & 23.31          \\
                             & FedProx & 29.79          & 46.80          & 44.67          & 35.12          & 39.09          &                              & FedProx & 16.48          & 17.66          & 29.83          & 27.98          & 22.99          \\
\multirow{3}{*}{CLIP}        & FedAVG  & 53.08          & 80.08          & 90.00          & 76.99          & 75.04          & \multirow{3}{*}{CLIP}        & FedAVG  & 65.60          & 57.64          & 71.64          & 75.42          & 67.57          \\
                             & FedProx & 66.06          & 87.33          & 91.68          & 78.42          & 80.87          &                              & FedProx & 65.60          & 57.64          & 71.64          & 75.42          & 67.57          \\
                             & Ours    & \textbf{96.34} & \textbf{97.91} & \textbf{99.76} & \textbf{85.59} & \textbf{94.90} &                              & Ours    & \textbf{78.00} & \textbf{63.69} & \textbf{87.52} & \textbf{87.79} & \textbf{79.25} \\ \bottomrule
\end{tabular}}
\end{table}

%% file: tab-person.tex
\begin{table}[t!]
\caption{Personalization accuracy. \textbf{Bold} means the best.}
\label{tab:my-table-person}
\resizebox{\textwidth}{!}{
\begin{tabular}{cllcccc|cllcccc}
\toprule
\multicolumn{1}{l}{Dataset} & \multicolumn{6}{c|}{PACS}                                                                                   & \multicolumn{7}{c}{Office-Home}                                                                                                         \\
\midrule
\multicolumn{1}{l}{Target}   & \multicolumn{1}{l}{BackBone} & Method  & C              & P              & S              & AVG            & \multicolumn{1}{l}{Target} & \multicolumn{1}{l}{BackBone} & Method  & C              & P              & R              & AVG            \\
\multirow{5}{*}{A}           & \multirow{2}{*}{AlexNet}     & FedAVG  & 72.86          & 61.08          & 78.22          & 70.72          & \multirow{5}{*}{A}         & \multirow{2}{*}{AlexNet}     & FedAVG  & 50.74          & 63.47          & 38.81          & 51.01          \\
                             &                              & FedProx & 71.37          & 56.89          & 81.53          & 69.93          &                            &                              & FedProx & 51.78          & 66.74          & 40.07          & 52.86          \\
                             & \multirow{3}{*}{CLIP}        & FedAVG  & 76.28          & 86.83          & 42.42          & 68.51          &                            & \multirow{3}{*}{CLIP}        & FedAVG  & 64.38          & 79.14          & 78.76          & 74.09          \\
                             &                              & FedProx & 90.81          & 90.42          & 63.95          & 81.73          &                            &                              & FedProx & 64.38          & 79.14          & 78.76          & 74.09          \\
                             &                              & Ours    & \textbf{97.65} & \textbf{99.40} & \textbf{86.75} & \textbf{94.60} &                            &                              & Ours    & \textbf{68.61} & \textbf{87.37} & \textbf{88.06} & \textbf{81.35} \\ \midrule
\multirow{6}{*}{C}           & \multicolumn{1}{l}{}         &         & A              & P              & S              & AVG            & \multirow{6}{*}{C}         & \multicolumn{1}{l}{}         &         & A              & P              & R              & AVG            \\
                             & \multirow{2}{*}{AlexNet}     & FedAVG  & 46.45          & 66.17          & 75.67          & 62.76          &                            & \multirow{2}{*}{AlexNet}     & FedAVG  & 23.51          & 61.78          & 41.56          & 42.28          \\
                             &                              & FedProx & 47.19          & 64.07          & 77.45          & 62.90          &                            &                              & FedProx & 24.54          & 64.04          & 40.18          & 42.92          \\
                             & \multirow{3}{*}{CLIP}        & FedAVG  & 84.11          & 92.81          & 81.02          & 85.98          &                            & \multirow{3}{*}{CLIP}        & FedAVG  & 73.81          & 80.38          & 80.48          & 78.23          \\
                             &                              & FedProx & 86.06          & 92.81          & 85.61          & 88.16          &                            &                              & FedProx & 73.81          & 80.38          & 80.48          & 78.23          \\
                             &                              & Ours    & \textbf{96.33} & \textbf{99.10} & \textbf{86.88} & \textbf{94.10} &                            &                              & Ours    & \textbf{78.97} & \textbf{87.60} & \textbf{87.60} & \textbf{84.72} \\ \midrule
\multirow{6}{*}{P}           & \multicolumn{1}{l}{}         &         & A              & C              & S              & AVG            & \multirow{6}{*}{R}         & \multicolumn{1}{l}{}         &         & A              & C              & R              & AVG            \\
                             & \multirow{2}{*}{AlexNet}     & FedAVG  & 37.65          & 75.00          & 81.53          & 64.73          &                            & \multirow{2}{*}{AlexNet}     & FedAVG  & 23.30          & 49.94          & 40.87          & 38.04          \\
                             &                              & FedProx & 35.45          & 73.93          & 83.57          & 64.32          &                            &                              & FedProx & 21.03          & 48.91          & 39.84          & 36.59          \\
                             & \multirow{3}{*}{CLIP}        & FedAVG  & 83.13          & 93.38          & 84.97          & 87.16          &                            & \multirow{3}{*}{CLIP}        & FedAVG  & 70.93          & \textbf{68.73} & 77.73          & 72.46          \\
                             &                              & FedProx & 83.86          & 93.59          & 88.54          & 88.66          &                            &                              & FedProx & 70.93          & \textbf{68.73} & 77.73          & 72.46          \\
                             &                              & Ours    & \textbf{97.56} & \textbf{97.65} & \textbf{86.75} & \textbf{93.99} &                            &                              & Ours    & \textbf{78.35} & 68.38          & \textbf{87.94} & \textbf{78.23} \\ \midrule
\multirow{6}{*}{S}           & \multicolumn{1}{l}{}         &         & A              & C              & P              & AVG            & \multirow{6}{*}{P}         & \multicolumn{1}{l}{}         &         & A              & C              & P              & AVG            \\
                             & \multirow{2}{*}{AlexNet}     & FedAVG  & 53.30          & 68.80          & 66.17          & 62.76          &                            & \multirow{2}{*}{AlexNet}     & FedAVG  & 22.27          & 49.14          & 58.51          & 43.31          \\
                             &                              & FedProx & 52.32          & 69.66          & 66.47          & 62.82          &                            &                              & FedProx & 20.21          & 50.06          & 58.29          & 42.85          \\
                             & \multirow{3}{*}{CLIP}        & FedAVG  & 90.71          & 94.02          & 94.91          & 93.21          &                            & \multirow{3}{*}{CLIP}        & FedAVG  & 69.07          & 66.21          & 77.79          & 71.02          \\
                             &                              & FedProx & 91.44          & 94.66          & 95.81          & 93.97          &                            &                              & FedProx & 69.07          & 66.21          & 77.79          & 71.02          \\
                             &                              & Ours    & \textbf{97.31} & \textbf{97.65} & \textbf{99.40} & \textbf{98.12} &                            &                              & Ours    & \textbf{78.56} & \textbf{68.50} & \textbf{87.37} & \textbf{78.14} \\ \bottomrule
\end{tabular}}
\end{table}

%% file: tab-comp.tex
\begin{table}[t!]
\caption{Comprehensive average accuracy. \textbf{Bold} means the best}
\label{tab:my-table-comp}
\resizebox{\textwidth}{!}{
\begin{tabular}{llllllllllll}
\toprule
Datasets & \multicolumn{5}{c}{PACS}                                        & \multicolumn{6}{c}{Office-Home}                                            \\
Backbone & \multicolumn{2}{c}{AlexNet} & \multicolumn{3}{c}{CLIP}          & Backbone & \multicolumn{2}{c}{AlexNet} & \multicolumn{3}{c}{CLIP}          \\
Methods  & FedAVG       & FedProx      & FedAVG & FedProx & Ours           & Methods  & FedAVG       & FedProx      & FedAVG & FedProx & Ours           \\
\midrule
A        & 60.93        & 59.89        & 64.65  & 77.81   & \textbf{95.04} & A        & 42.18        & 43.77        & 71.97  & 71.97   & \textbf{80.51} \\
C        & 57.99        & 58.88        & 84.50  & 87.95   & \textbf{95.06} & C        & 35.96        & 36.60        & 73.08  & 73.08   & \textbf{79.46} \\
P        & 59.68        & 59.41        & 87.87  & 89.42   & \textbf{95.43} & P        & 36.42        & 34.90        & 72.26  & 72.26   & \textbf{80.55} \\
S        & 56.14        & 55.89        & 89.16  & 90.08   & \textbf{94.99} & R        & 39.73        & 39.13        & 72.12  & 72.12   & \textbf{80.55} \\
AVG      & 58.69        & 58.52        & 81.55  & 86.32   & \textbf{95.13} & AVG      & 38.57        & 38.60        & 72.36  & 72.36   & \textbf{80.27}
\\ \bottomrule
\end{tabular}}
\end{table}

%% file: tab-vlcs.tex
\begin{table}[t!]
\caption{Comprehensive average accuracy on VLCS. \textbf{Bold} means the best}
\label{tab:my-table-comp-vlcs}
\centering
\resizebox{0.5\textwidth}{!}{
\begin{tabular}{llllll}\toprule
Backbone & \multicolumn{2}{c}{AlexNet} & \multicolumn{3}{c}{CLIP}          \\
Methods  & FedAVG       & FedProx      & FedAVG & FedProx & Ours           \\ \midrule
C        & 62.13        & 61.37        & 72.48  & 68.57   & \textbf{83.68} \\
L        & 63.01        & 63.77        & 75.04  & 76.50   & \textbf{82.62} \\
S        & 63.15        & 63.59        & 68.13  & 75.50   & \textbf{82.82} \\
V        & 62.32        & 62.04        & 69.55  & 70.09   & \textbf{83.30} \\
AVG      & 62.65        & 62.69        & 71.30  & 72.67   & \textbf{83.11} \\ \bottomrule
\end{tabular}}
\end{table}